# Desarrollo de un Robot de Rehabilitación pasiva para la articulación de la muñeca mediante la implementación de un microcontrolador Arduino UNO

## Development of a passive Rehabilitation Robot for the wrist joint through the implementation of an Arduino UNO microcontroller

E. A. Ceballos [1], M. Díaz-Rodriguez [2], J. L. Paredes [3], P. C. Vargas

[1] Grupo de investigación del Laboratorio de Mecatronica y Robotica, Dpto. de Tecnologia para el Diseño Indsutrial, Universidad de Los Andes, Venezuela. Email: eceballos@ula.ve
[2] Grupo de investigación del Laboratorio de Mecatronica y Robotica, Dpto. de Tecnologia y Diseño. Universidad de Los Andes, Venezuela. Email: dmiguel@ula.ve
[3] Grupo de investigación GIBULA, Dpto. de Circuitos y Medidas, Universidad de Los Andes, Venezuela. Email: paredesj@ula.ve

**Resumen**

En esta investigación se implementó el uso de un microcontrolador Arduino UNO R3 para controlar los movimientos de un prototipo funcional robótico desarrollado para realizar ejercicios de rehabilitación en la articulación de la muñeca; este dispositivo puede ser empleado para asistir al médico fisiatra para rehabilitar los cuadros de tendinitis, sinovitis, artritis reumatoidea y para procesos de terapia pre y post operatoria en dicha articulación. Durante la etapa de diseño del prototipo funcional, se utilizó la metodología del *proceso de diseño industrial desde enfoque de la ingeniería concurrente*, mediante la cual se pudieron realizar estudios antropométricos relacionado con las dimensiones y ángulos de movimiento de la articulación de la muñeca en la población venezolana. A partir de la información recolectada se elaboró la propuesta de diseño, y con la utilización de programas CAD se definieron las diferentes formas, geometrías y materiales de los componentes del dispositivo de rehabilitación, que posteriormente fueron analizados mediante el método de elementos finitos para la determinación del estado tensional de esfuerzos y factores de seguridad mediante la utilización de programas CAE. Adicionalmente se desarrolló un software para la adquisición, registro, reproducción y ejecución de los diferentes movimientos producidos durante la realización de la terapia de rehabilitación. Mediante la investigación desarrollada se logró diseñar un dispositivo que ayudará a la rehabilitación de la articulación de la muñeca permitiendo realizar la combinación de movimientos de flexión dorsal-palmar y cubital-radial para la recuperación de la funcionalidad de la articulación de diversas patologías presentadas en la población venezolana.

**Palabras Clave:** Rehabilitación de articulación de la muñeca; Dispositivo de registro y ejecutor; diseño; microcontrolador arduino; robótica.

**Abstract**

In this research was implemented the use of an Arduino UNO R3 microcontroller to control the movements of a prototype robotic functional developed to perform rehabilitation exercises in the wrist joint; This device can be used to assist the physiatrist to rehabilitate the tendinitis, synovitis, rheumatoid arthritis and for pre-operative and post-operative therapy in this joint. During the design stage of the functional prototype, the methodology of the industrial design process was used from a concurrent engineering approach, through which anthropometric studies could be performed related to the dimensions and angles of movement of the wrist joint in the population Venezuelan From the information collected, the design proposal was elaborated, and the use of CAD programs defined the different forms, geometries and materials of the components of the rehabilitation device, which were later analyzed using the finite element method for the determination The tensional state of efforts and safety factors through the use of CAE programs. In addition, a software was developed for the acquisition, registration, reproduction and execution of the different movements produced during the rehabilitation therapy. Through the research developed, a device was designed that will help the rehabilitation of the wrist joint allowing the combination of dorsal-palmar flexion and ulnar-radial movements to recover the joint function of various pathologies presented in the Venezuelan population.

## 1. INTRODUCCIÓN.

Debido a la gran cantidad de pacientes que ingresan a los servicios de Fisiatría de los Hospitales Venezolanos con lesiones en las articulaciones superiores, pudiendo estas haber sido ocasionadas por diversas causas (impactos o mal uso de la articulación), estas lesiones han originado una gran demanda de personal capacitado para atender las diversas patologías en dichos servicios. Por otra parte, los especialistas en esta área mencionan que existen adicionalmente otras causas de lesiones en dichas articulaciones relacionadas con su disfuncionalidad, esta puede deberse a la degeneración de los tejidos de las articulaciones, la cual puede afectar significativamente la movilidad de dicha articulación, pudiendo llegar al extremo de incapacitar por completo la movilidad de la articulación, la cual pudiera ir acompañada con severas inflamaciones en los tejidos de los músculos y tendones. Otras de las patologías recurrentes al servicio de fisiatría corresponde a los casos de tendinitis y tendinopatías crónicas mostradas en el *Journal of American Academy of Orthopedic Surgery* [1]. Dentro de la gama de tratamientos médicos dirigidos para la rehabilitación dichas afecciones en la articulación de la muñeca, se encuentran definidas una series de etapas de rehabilitación propuestas en el Handbook Of Orthopaedic Rehabilitation de 1996 [2]; en este material se encuentra planteado el uso de un protocolo de rehabilitación desarrollado en tres etapas principales. Una primera etapa que consiste con la inmovilización de la articulación, una segunda etapa de exposición de los tendones y músculos a cambios térmicos entre frío y calor, y una tercera etapa en la que el médico fisiatra realiza movimientos suaves en la articulación para ir recuperando la tonificación y la masa muscular [2]. Motivado a que en la segunda y en la tercera etapa el médico fisiatra se encarga de ejercitar dichos músculos durante largos periodos de tiempo resultando una actividad fatigante, se planteó la necesidad de diseñar una estación de rehabilitación que permita co-ayudar a ejecutar dichos movimientos programados y controlados. Mediante el diseño de este dispositivo se busca facilitar la labor del médico fisiatra en la rehabilitación de la articulación de la muñeca.

Otros de los aspectos importante a considerar para el planteamiento de este trabajo, es el hecho que la gran mayoría de los equipos utilizados en terapias de rehabilitación se caracterizan por poseer diseños estructurales exoesqueléticos con atributos de escasa portabilidad y movilidad, niveles altos de consumo energético y gran volumen, por lo que ocupan considerable espacio en las salas de rehabilitación. Al considerar estos aspectos, en este trabajo, se plantea el desarrollo de un robot de rehabilitación controlado a través de un microcontrolador Arduino Uno R3; que mediante la implementación de dicho controlador se podrá registrar y ejecutar las diferentes rutinas de rehabilitación para las diversas afecciones en la articulación de la muñeca de forma automatizada y controlada. Otro aspecto relevante en el desarrollo del dispositivo consiste en la reducción del volumen ocupado por este dentro de la salas de rehabilitación.

Esta investigación se encuentra estructurada de la siguiente manera: en el apartado número dos se presenta el marco teórico, en el cual se muestran las investigaciones previas relacionadas con el desarrollo del dispositivo de rehabilitación. En el apartado número tres se presenta la metodología de diseño utilizada para el diseño del dispositivo de rehabilitación, junto con la especificación de los requerimientos de diseños necesarios para la definición del prototipo. En el apartado número cuatro se presentan los análisis de ingeniería realizados al prototipo propuesto y por último en el apartado número cinco, se presentan las conclusiones finales extraídas de la investigación realizada.

## 2. ANTECEDENTES.

Gracias a los grandes avances tecnológicos alcanzados en el área de robótica y en las ciencias medicas de rehabilitación, diversas universidades y centros de investigaciones han desarrollado variados prototipos robóticos destinados a asistir a los médicos fisiatras en la ejecución de ejercicios de rehabilitación para las articulaciones de los miembros superiores e inferiores. Dentro de la gama de equipos utilizados para la rehabilitación de la articulación de la muñeca, encontramos equipos como el Haptic Robot, el cual es un dispositivo robótico desarrollado por el Instituto para la Rehabilitación de Eslovenia que se basa en un diseño de armazón estructural tipo exoesqueleto con fijación al suelo; este dispositivo permite la rehabilitación del brazo, antebrazo y muñeca en un rango de movilidad significativo. Adicionalmente se caracteriza por poseer un controlador de movimiento activado por un actuador lineal acoplado sobre una rótula esférica [3]. Entre otro de los dispositivos destacados en rehabilitación de muñeca y antebrazo, es el desarrollado por el Instituto Italiano di Tecnología [4]. Este robot posee tres grados de libertad controlados por tres motores rotatorios DC acoplados con un sistema de transmisión de engranaje cónico, el cual cuenta con un protocolo terapéutico que permite la restauración de la funcionalidad de la muñeca en pacientes con accidente cerebrovascular crónico. A principios del 2007 Investigadores del Instituto Tecnológico de Massachusetts (MIT) desarrollaron un robot de terapia para la rotación de la muñeca. Este dispositivo exoesqueletico consta de 3 grados de libertad, lo cual permite ejecutar los ejercicios de rehabilitación con un mayor control con respecto a sus antecesores [5]. A mediados del 2007, investigadores de la Universidad Militar Nueva Granada desarrollaron un robot exoesqueletico para la rehabilitación motora de la articulación del hombro y del antebrazo [6]. Este sistema presenta dos grados de libertad accionados por un conjunto de motores paso a paso con sistemas de transmisión de engranes, lo cual le atribuye una mayor movilidad.

A finales del 2008, Investigadores de la Universidad Rice en Houston, U.S.A, desarrollaron un dispositivo de rehabilitación exoesquelético con fines de rehabilitación y entrenamiento para pacientes con lesiones neurológicas en las articulaciones de la muñeca y antebrazo, [7]. Este dispositivo se caracteriza por ser un robot de cuatro grados de libertad accionados por motores rotatorios DC acoplados a sistemas de transmisión de cabrestante; en dicho diseño se observa la carencia de portabilidad. En el año 2013, investigadores de la Universidad de Hong Kong [8], desarrollaron un robot de rehabilitación motora controlado a partir de la detección y filtrado de las señales mioeléctricas, la cuales le permitía al paciente ejecutar el movimiento controlado y preciso al detectar las diferentes de señales mioeléctricas capturadas en el antebrazo. Adicionalmente a finales del 2013 investigadores de la Universidad del Quindío y de la Universidad del Valle de Colombia, desarrollaron una propuesta de framework multi-nivel empleadas en laboratorios de experimentación de robot seriales de acceso remoto, en los cuales permiten controlar el posicionamiento y las trayectorias de robots desde un laboratorio que se encuentre distante a la ubicación del robot a ser controlado [9]. Con esta propuesta han experimentado con robots de cinco grados de libertad como el robot Mitsubishi RV-2A, en donde el calculo del posicionamiento y la dinámica del movimiento son modeladas a través de técnicas y principios matemáticos presentados en [10]. Es importante destacar que este tipo de propuesta abre un gran campo de acción para el desarrollo de robot de rehabilitación manejados de forma remota, facilitando a un gran avance al campo de la telemedicina de rehabilitación. En resumen se puede apreciar que la mayoría de equipos diseñados para la rehabilitación de muñeca son dispositivos exoesqueléticos utilizados como estaciones de rehabilitación, es decir, el antebrazo y la muñeca del paciente se apoyan directamente sobre el equipo de rehabilitación. Otros de los aspectos relevantes de estos equipos es que se encuentran diseñados para moverse dentro del rango de movimiento de la articulación de la muñeca y adicionalmente se encuentran activados por motores lineales o rotacionales de paso-paso.

## 3. METODOLOGÍA.

Al inicio de este apartado se presenta la metodología de diseño utilizada para el desarrollo del dispositivo de rehabilitación, basada en la metodología del proceso de Diseño Industrial planteada por [11]. Mediante el empleo de esta metodología se establecieron siete fases de diseño. Fase 1: Identificación de las necesidades. En esta fase se identificaron las necesidades a satisfacer para el desarrollo del equipo de rehabilitación de muñeca para pacientes con medidas antropométricas venezolanas, junto con la definición del usuario y el contexto. Fase 2: Especificación de los requerimientos de diseño. Dentro de esta fase se definieron los principales requerimientos de diseño, tales como los Funcionales, Ergonómicos, Tecnológicos y Formales. Fase 3: Diseño Conceptual del Dispositivo de Rehabilitación. En esta etapa de diseño, se generaron variados bocetos del equipo de rehabilitación para representar a través de estos, la forma conceptual de dicho equipo tomando en cuenta los diferentes requerimientos de diseño planteados en la fase 2. Fase 4: Diseño Preliminar e Ingeniería Básica. En esta etapa de diseño se generó el dimensionamiento y la forma de los diferentes componentes que constituirán el equipo de rehabilitación de muñeca a partir de las dimensiones antropométricas consideradas en las fases anteriores. Fase 5: Diseño final e Ingeniería de detalle. En esta fase realizaron los análisis cinemático, cinéticos y de esfuerzos relacionados con el comportamiento del dispositivo antes las cargas generadas por el peso de la mano. Fase 6: Desarrollo del Programa Controlador. En esta fase se desarrolló un programa de interface para la adquisición de datos y control de los motores actuadores en la plataforma Matlab GUI. Mediante este programa

E.A. Ceballos, M. Díaz-Rodríguez, J.L. Paredes,
P.C. Vargas3, Autor 4

se estableció la interface con el microcontrolador Arduino Uno R3 para la adquisición de la data y control de los motores actuadores. Fase 7: Construcción del Prototipo Conceptual del Dispositivo de Rehabilitación. En esta última fase se construyó un dispositivo de prueba conceptual con el fin de evaluar el funcionamiento del software controlador y mostrar la interacción del dispositivo con el usuario. El ensamblado del prototipo conceptual se realizo a partir de la unión de diferentes componentes comerciales empleados para otros fines y junto a la incorporación de piezas fabricadas.

### 3.1 Identificación de las necesidades y requerimientos de diseño.

En este apartado se establecieron los aspectos de diseño básicos para satisfacer las necesidades plasmadas dentro del planteamiento del problema presentado en el primer apartado, definiendo de esta manera las características principales del paciente (usuario del equipo de rehabilitación), junto al entorno donde se plantea el funcionamiento del equipo.

**3.1.1 Definición del usuario y del contexto geográfico, la definición del usuario (paciente)**. En esta fase se determinó las características principales del paciente. Este usuario se caracteriza por pertenecer a un grupo de pacientes adolescente-adulto que presentan patologías como tendinitis, artritis reumatoide, sinovitis, rehabilitación post-operatoria y que recurren al Servicio de Fisiatría del Hospital Universitario IAHULA de la ciudad de Mérida, con una edad comprendida entre los 15 a 50 años y con un diagnostico de nutrición normal (percentil 50). Para este percentil, el peso del adolescente-adulto varía entre los 55 kg a 90 kg y presentan una altura oscilante desde los 125 cm hasta los 177,5 cm [12].

**3.1.2 Definición de las medidas antropométricas de la articulación de la muñeca**. Al realizar un análisis antropométrico de la mano y del antebrazo, se obtuvieron los siguientes datos: el peso del antebrazo y mano corresponde en promedio al 2,2% del peso total del paciente, lo cual se aproxima a 1,5 kg. Distancia del centro de masa de la mano a la articulación de la muñeca 7,11 cm [13] peso de la mano promedio 0,5 kg. Según la norma DIN 33 402.2°, el ancho de la articulación de la muñeca para un percentil 95 se ubica alrededor 18,9 cm, al igual que el perímetro de la mano (19,5 cm P5%; 21,0 cm P50%; 22,9 cm P95%) [14], serán datos importantes que se consideraron para definir la geometría y forma del dispositivo de rehabilitación para la articulación de la muñeca.

**3.1.3 Especificación de los requerimientos de diseño (Fase 2)**. Dentro de los principales requerimientos de diseño, se establecieron los siguientes: a) Requerimientos funcionales; garantizar la estabilidad del miembro mientras se realiza el movimiento, permitir el movimiento de arco dentro de un rango de -50° hasta +50° en flexión dorsal-palmar y un rango de -15° a +15° en flexión cubital-radial debido a que en la segunda fase de rehabilitación se deben realizar movimientos de arco limitados para evitar posteriores lesiones [2], soportar a nivel estático los esfuerzos y deformaciones del sistema y soportar las cargas dinámicas del sistema, y poseer la capacidad para registrar, almacenar y ejecutar la rutina de rehabilitación establecida por el especialista. b) Requerimientos ergonómicos: adaptarse a las dimensiones del antebrazo y de la mano para percentiles 5, 50 y 95. Instalación fácil, rápida y entendible, permitir la movilización suave y controlada sin generar un dolor intenso sobre el paciente y poseer diferentes puntos de sujeción para la estabilización de la mano. c) Requerimientos formales. Usar geometrías curvas que permitan mantener una armonía integral estética entre el dispositivo y el miembro en tratamiento, y permitir que la carcasa y otros componentes puedan ser pintados dentro de una gama de colores para poder personalizar el dispositivo para ambos sexo. d) Requerimientos tecnológicos. Ser construido con materiales disponible en el país, emplear procesos de manufactura sencillos y económicos que puedan ser utilizados con tornos, fresadoras, centros de mecanizado de control numérico, entre otras máquinas de mecanizado disponibles en el país, y ser fabricable con mano de obra nacional y poseer repuestos de los componentes electrónicos en el país.

**3.1.4 Diseño conceptual del equipo de rehabilitación de la muñeca (Fase 3).** En esta etapa de diseño, se generó una descripción formal del equipo de rehabilitación a través de bocetos sencillos; tomando en cuenta los diferentes requerimientos de diseño planteados en el apartado anterior, partiendo del concepto que el dispositivo debe permitir la movilidad y la rotación en dos ejes principalmente, debe brindar la estabilización en la articulación de la muñeca y adicionalmente debe ser capaz de adaptarse a la anatomía de diferentes tamaños de manos.

### 3.2 Diseño preliminar e ingeniería básica de las propuestas del equipo rehabilitador de muñeca (Fase 4).

En esta etapa del diseño preliminar se utilizaron las medidas antropométricas recopiladas en las fases 1 y 2, para generar el dimensionamiento y la forma de los diferentes componentes que formaran parte de la propuesta final del equipo de rehabilitación de muñeca.

**3.2.1 Definición de materiales a emplear en el dispositivo**. Se emplearon para el desarrollo y elaboración del dispositivo de rehabilitación, aleaciones de aluminio 6061 (Esfuerzo Max. Tracción 95 MPa), acero aleado de alta resistencia (Esfuerzo Max. Tracción 210 MPa) debido a su fácil acceso en el país, Plástico Acrilonitrilo Butadieno Estireno (ABS) (Esfuerzo Max. Tracción 27,6 MPa - 55,2 MPa), [15], [16].

**3.2.2 Definición esquemática de la propuesta de diseño del dispositivo de rehabilitación**. En este apartado se presenta de forma detallada la descripción y principales características de la propuesta presentada para el desarrollo del equipo de rehabilitación de muñeca. Se desarrollaron bocetos basados en cadenas cinemáticas cerradas [17] así como también propuesta como un dispositivo robótico serial esférico estacionario [18]. Se desarrollaron propuestas principalmente de dos grados de libertad, permitiendo la rotación en el plano flexión-extensión Palmar-Dorsal y en el plano Cubital-Radial de la articulación de la muñeca. Este dispositivo se caracteriza por ser usado como una estación de rehabilitación teniendo la capacidad de grabar, almacenar y ejecutar rutina pre-grabadas y en vivo realizadas por el médico especialista, este dispositivo cuenta con motores paso a paso y encoders rotacionales (STM17S-1X Simple), adicionalmente cuenta un sistema de transmisión por engranes con relación 4:1, que permite una mayor precisión del movimiento y mayor potencia de trabajo. Este dispositivo se encuentra constituido principalmente por dieciocho piezas mecánicas, como se aprecia en la figura 1.

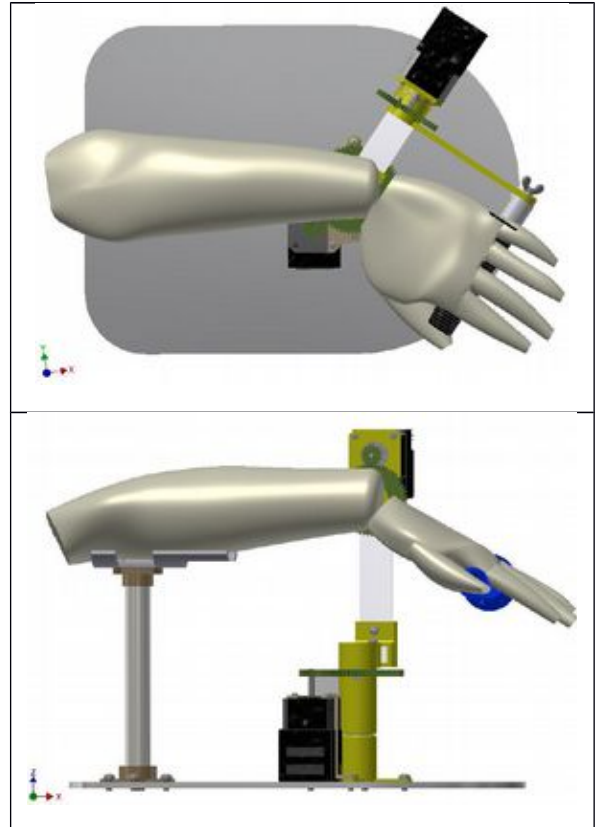

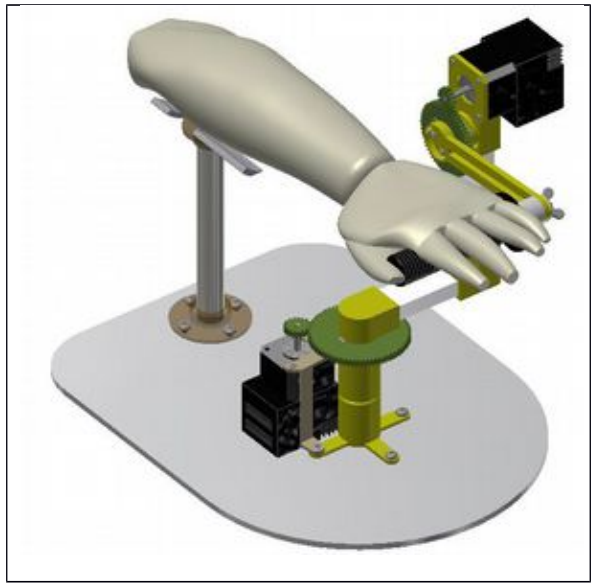

**Figura 1.** Propuesta de diseño del dispositivo de rehabilitación. Descripción de (a) vista isométrica (b) vista superior con movilidad en plano radial-cubital (c) vista lateral con movilidad en plano dorsal-palmar. Elaboración propia

**3.3 Desarrollo del prototipo funcional-conceptual del dispositivo de rehabilitación (Fases 6 y 7).**

El prototipo funcional-conceptual para rehabilitación de muñeca se desarrolló con el fin de probar el concepto de diseño y adicionalmente para obtener, almacenar y ejecutar los datos obtenidos de las rutinas reales realizadas por los especialistas de fisiatría en el Servicio de Fisiatría del Hospital Universitario de los Andes IAHULA. El dispositivo funcional-conceptual consta de una superficie de apoyo para el antebrazo, un elemento de sujeción para pacientes con capacidad de sujeción de la mano y un elemento o accesorio para pacientes con discapacidad en sujeción de la mano, como se muestra en la figura 2, adicionalmente al dispositivo se le adaptó una base de cámara móvil modelo PT785-S activada por servos motores HITEC HS-785HB desarrollada por la compañía Servocity [19]. A esta base de cámara se le adaptó un elemento que hace la función de mango de agarre, siendo esta pieza impresa en 3D en plástico Poli-Acido Làctico (PLA). Otros de los elementos adaptados al dispositivo fuerón los dos potenciómetros rotacionales, los cuales fuerón instalados en los ejes de rotación de la base de la cámara. Estos elementos permiten sensar la posición angular del dispositivo en un rango de 270°, que luego son adquiridos mediante la utilización del microcontrolador ARDUINO UNO R3. Este microcontrolador es una plataforma electrónica open-hardware que se emplea para la creación y diseños de prototipos, y que adicionalmente permite conectar sensores, actuadores y otros elementos mediante sus entradas y salidas, analógicas y digitales.

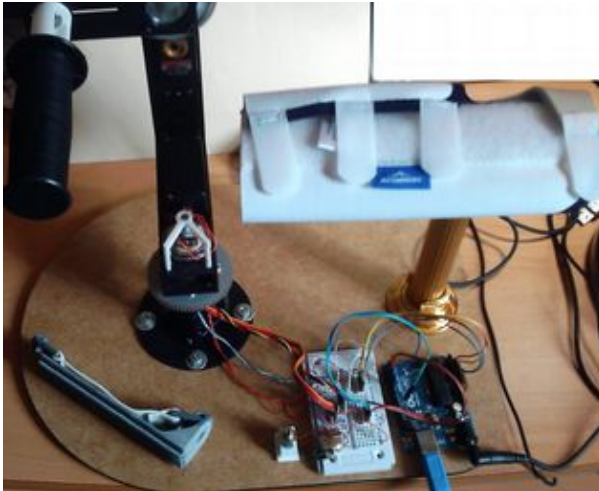

**Figura 2.** Dispositivo funcional-conceptual del equipo de rehabilitación. Elaboración propia

El programa desarrollado se encuentra conformado principalmente por variados módulos que controlan las diversas funciones del dispositivo de rehabilitación. Como se aprecia en la figura 3. Dentro de estos módulos básicos se destacan los siguientes: a) Módulo de conexión, (se utiliza para la conexión entre el programa controlador y la tarjeta Arduino UNO R3. Al activar el botón inicia el proceso de conexión entre ambos elementos). b) Módulo de control de servos manual, (permite controlar cada grado de libertad del dispositivo de forma independiente mediante barras de comando slider. Adicionalmente, permite posicionar o ajustar manualmente la posición inicial de trabajo del dispositivo). c) Modulo temporizador, (permite contabilizar el tiempo transcurrido durante la realización de la rutina de rehabilitación). d) Módulo de adquisición de datos, (permite grabar las diversas rutinas de rehabilitación en un tiempo determinado de forma discreta). e) Módulo de ejecución de rutinas de rehabilitación pre-registradas, (permite ejecutar las rutinas grabadas temporalmente en el módulo de adquisición de data). f) Módulo de parada de emergencia, (permite parar la ejecución de las rutinas de forma inmediata en caso que se presente algún imprevisto). g) Módulo de monitoreo, (permite monitorear en tiempo real las rutinas pregrabadas o grabadas al instante, así como monitorear en tiempo real la ejecución de las rutinas pregrabadas, cabe destacar que en este módulo se observa la variación de la posición angular del dispositivo en los dos planos de estudio con respecto al tiempo). h) Módulo de almacenamiento temporal, (se almacena temporalmente los datos adquiridos de posicionamiento angular de los dos planos de estudio del dispositivo). i) Módulo de almacenamiento y ejecución permanente, (a través de este módulo el programa permite almacenar, cargar y ejecutar las rutinas pregrabadas o grabadas al momento mediante el uso de archivos de datos de formato .xls).

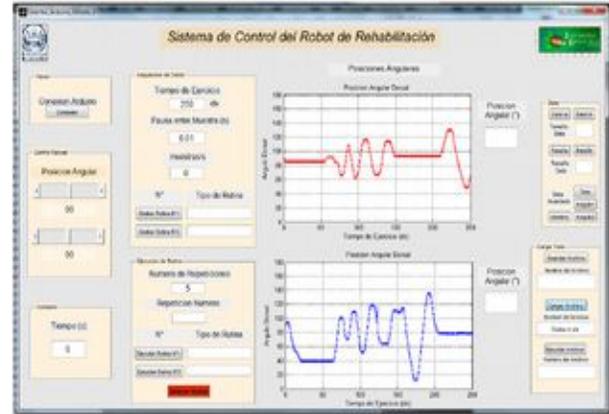

**Figura 3.** Interfaz de control desarrollada entre MATLAB GUI y el software Arduino 1.0.6. Elaboración propia

## 4. DISCUSIÓN Y RESULTADOS.

*Adquisición de los datos reales del dispositivo de rehabilitación.* A través del dispositivo funcional-conceptual de rehabilitación se pudo adquirir rutinas de rehabilitación de la articulación muñeca en tiempo real realizadas por los médicos especialistas del servicio de fisiatría del IAHULA para la comprobación del funcionamiento del mismo, pudiéndose obtener así datos reales para realizar los análisis cinemáticos y cinéticos en la propuesta final presentada en el apartado anterior. Al conocer los datos de posicionamiento angular con respecto al tiempo se pudo analizar de forma más detallada el comportamiento estructural y cinemático del dispositivo de rehabilitación en ambiente controlado. La rutina real, es una rutina de movimiento circular combinado en el que se inicia con movimientos de arco circulares limitados y se va incrementando el tamaño del arco circular con respecto al tiempo tanto en el plano lateral como en el plano flexión-extensión dorsal-plamar, ver figura 4.

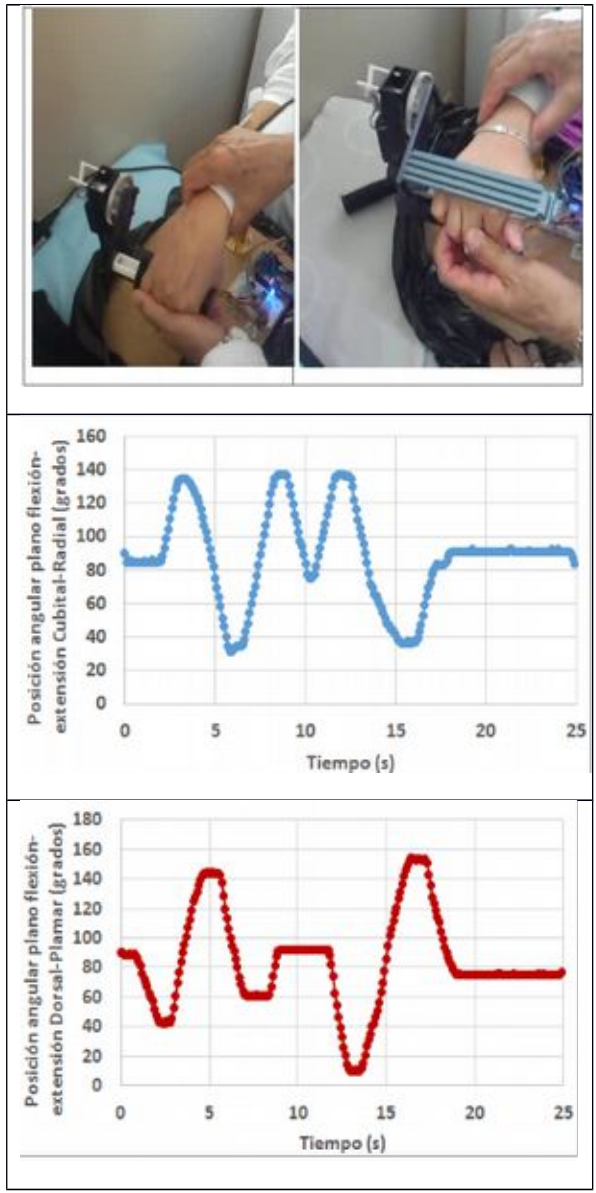

**Figura 4.** Rutina de rehabilitación realizada por los especialistas del IAHULA. Elaboración propia.

### 4.1 Análisis de ingeniería de detalle realizados al equipo rehabilitación de muñeca (Fase5).

A partir de la propuesta generada en la fase anterior, se procedió a realizar análisis cinemático, cinéticos y de es-fuerzos relacionados con el comportamiento del dispositivo ante las cargas generadas por el peso de la mano, este tipo de análisis forma parte de la fase de ingeniería de detalle, en donde se determina el campo tensional, deformación y factor de seguridad de los componentes que definen el equipo rehabilitación de muñeca.

**4.1.1 Análisis biomecánico**, en este análisis se estimó de forma teórica la carga a vencer por el dispositivo de rehabilitación en forma pasiva, es decir, en este caso el paciente no ofrece resistencia al movimiento de la articulación de la muñeca, si no que más bien permite que el dispositivo ejercite la articulación. Esto es motivado debido a que el dispositivo se encuentra concebido a ayudar al médico fisiatra en la ejecución de las rutinas durante la segunda fase de rehabilitación, en la cual se requieren de movimientos suaves de arcos limitados para la recuperación de la movilidad de la articulación de la muñeca. Al considerar el peso de la mano aproximado de 0,5 kg para una persona joven entre percentil 50 a 95, y considerando que la distancia aproximada del centro de gravedad de la mano a la articulación de la muñeca se encuentra alrededor de los 7,11 cm [13], [20], es posible entonces estimar el momento mínimo necesario a vencer por el dispositivo de rehabilitación. Este momento se puede determinar aplicando la ecuación (1).

$$M_{min} = W_m \cdot L_m \qquad (1)$$

Donde:
$M_{min}$ : Momento minimo;
$W_m$ : Peso de la mano;
$L_m$ : Distancia del centro de gravedad de la mano al centro de rotación de la articulación.

Al sustituir los valores en ecuación (1), se obtiene el siguiente resultado:

$$M_{min} = \left(0,5\,kg \cdot 9,81\,\frac{m}{s^2}\right) \cdot (0,0711\,m) = 0,3487\,N \cdot m$$

$$M_{min} = 348,7\,N \cdot mm$$

Este valor indica que los actuadores rotacionales deben poseer un par motor mayor a 0,3487 N·m para poder mover la articulación de la muñeca de forma pasiva.

**4.1.2 Análisis cinemático y cinético**. En este análisis cinemático-cinético se estudió el comportamiento de las variaciones de la posición, velocidad y momento impulsor del dispositivo de rehabilitación, a partir de las rutinas reales de rehabilitación de muñeca grabadas con el dispositivo funcional-conceptual mostrado en el apartado anterior. La información obtenida a través del dispositivo de rehabilitación funcional-conceptual se modeló en el módulo cinemático del software Autodesk Inventor 2015. En este software se definieron las variaciones de posición angular de los planos flexión-extensión dorsal-palmar y cubital-radial en los pares cinemáticos de revolución ubicados en los motores paso-paso.

**4.1.3 Análisis de posición angular**. Al definir las variaciones de las posiciones angulares con respecto al tiempo en los pares de revolución cinemático de los motores paso-paso tomadas de las rutinas reales presentadas en el apartado anterior, se obtuvieron los siguientes resultados mostrados en la figura 5.

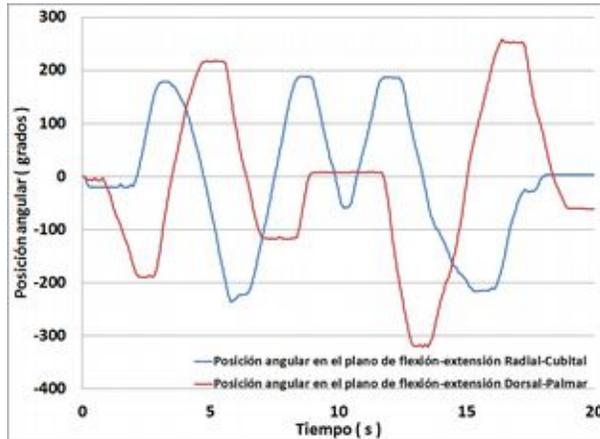

**Figura 5.** Variación de la posición angular Vs. tiempo en los planos Dorsal-Palmar y Cubital-Radial. Elaboración propia.

En la figura 5, se puede apreciar la variación de la posición angular en los pines de los motores paso-paso con respecto al tiempo, esta variación de la posición angular se mantiene en los motores paso-paso dentro de un rango de -310° a +260° en el plano de flexión-extensión y para el plano lateral el rango de rotación se ubica entre los -240° a 190° de rotación. Puede observarse adicionalmente que el ángulo máximo de rotación corresponde al valor de -315,5°. Es importante destacar que estos ángulos presentan una relación 4:1 con respecto a la posición original de la mano, esto debido a la transmisión de engranes de 4:1 diseñada.

**4.1.4 Análisis de velocidad angular**. En este análisis se estudiaron la variación de las velocidades angulares en los pares cinemáticos de revolución correspondiente a la ubicación de los motores paso-paso en los diferentes planos de estudios, como se aprecia en la figura 6.

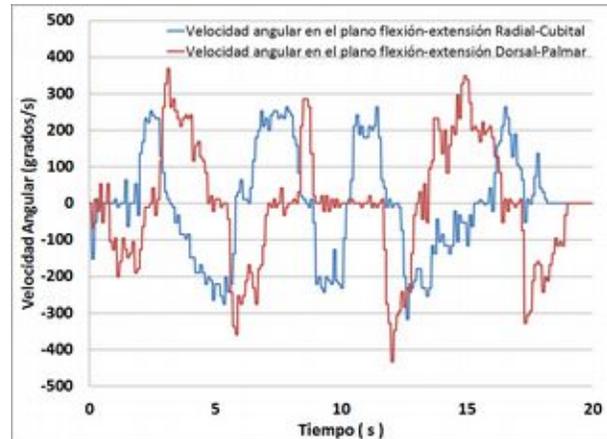

**Figura 6.** Variación de la velocidad angular Vs. tiempo en los planos flexión-extensión Dorsal-Palmar y Cubital-Radial. Elaboración propia.

En la figura 6, se puede apreciar la variación de la velocidad angular del dispositivo en los planos flexión-extensión Dorsal-Palmar y Cubital-Radial, cabe destacar que se alcanzó medir una velocidad angular máxima ins-tantánea de 432,42 °/s en el plano dorsal-palmar, cuyo valor se tomó en cuenta para la selección de final de los motores paso-paso que se utilizaran en el dispositivo de rehabilitación final.

**4.1.5 Análisis de momento impulsor**. En este análisis se presenta la variación del momento impulsor de los motores paso-paso en los diferentes planos de estudios, como se muestra en la figura 7.

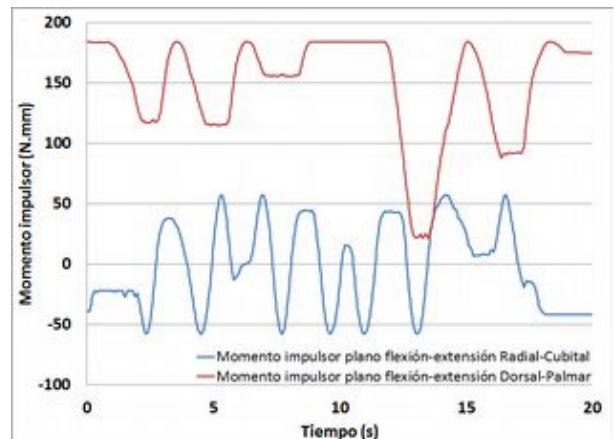

**Figura 7.** Variación del momento impulsor Vs. tiempo en los pares cinemáticos de los motores paso-paso. Elaboración propia.

En la figura 7 se puede observar la variación del momento impulsor con respecto al tiempo, en el caso del plano lateral el momento impulsor varía en un rango desde -60 N·mm hasta +60 N·mm, en el caso del plano de flexión-extensión Dorsal-Palmar el momento impulsor varía desde 40 N·mm hasta un máximo de 184,5 N·mm. este valor máximo se debe al peso de la mano y del mismo dispositivo que debe vencer el motor paso-paso en el plano de flexión-extensión Dorsal-Palmar para alcanzar el movimiento deseado. Cabe mencionar que estos valores obtenidos a través de la simulación dinamica, son cuatro veces menores a los valores del momento impulsor calculados en los ejes de rotación del dispositivo. Esto es debido a los sistemas de transmisión por engranes 4:1 planteados en el diseño [21]. De este análisis se puede concluir que se requiere de un motor paso-paso que posea un par motor mayor a los 184,5 N·mm para lograr ejercer movimiento del dispositivo alrededor de dicho plano, con este dato se podrá realizar la selección de los motores paso-paso de manera más objetiva.

*Selección de los motores actuadores*, de los análisis cinemáticos y cinéticos previos se pudo obtener los siguientes valores de trabajo como los son: las velocidades angulares instantánea máximas del dispositivo en los planos de rotación se encuentran alrededor de los 432,42 °/s, y el valor del par mínimo necesario para lograr los movimientos de 184,5 N·mm, al considerar estos valores se optó por seleccionar el motor paso-paso STM17, el cual se encuentra conformado con un sistema encoder que permitirá detectar la posición del dispositivo en cualquier instante de tiempo. Este modelo corresponde a un modelo de motor paso-paso de la serie NEMA 17, el cual puede alcanzar una velocidad máxima de 3600 °/s al conectarse a un controlador simple [22] o incluso uno avanzado si se quieren tomar en cuenta aspectos de control de fuerza [23], permitiendo generar un par de trabajo mayor a 480,18 N·mm que es adecuado para las rutinas de rehabilitación al dispositivo.

**4.1.6 Análisis estructural**. En este apartado se realizó un análisis estructural estático de los sistemas y componentes que conforman al equipo, en este se estudió los efectos que producen la carga producida por el peso de la mano (0,5 kg) y el peso del antebrazo (1,5 kg) en la propuesta final seleccionada [13]. *El análisis estático del cuerpo estructural del equipo*, se desarrolló en el software Autodesk INVENTOR 2015, mediante la aplicación de tres subprocesos principales: a) Pre-procesador. En este subproceso se definieron las cargas, las condiciones de apoyos, materiales y definición de las principales características del generado de mallado del modelo, b) Procesador. Se generan las matrices de rigidez y desplazamiento para los nodos y elementos que constituyen el mallado del modelo, c) Post-Procesador. Se extrajeron el valor de los esfuerzos de Von Mises y los factores de seguridad de los componentes que conforman al equipo. Para el caso de los esfuerzos de Von Mises se encontraron que las piezas críticas resultaron ser la arandela y el tornillo de ajuste mariposa alcanzado el valor de esfuerzo máximo de 159,41 MPa, y adicionalmente se obtuvieron valores de factores de seguridad superiores a 1,5, indicando que no presenta falla por deformación plástica [16], ver figura 8.

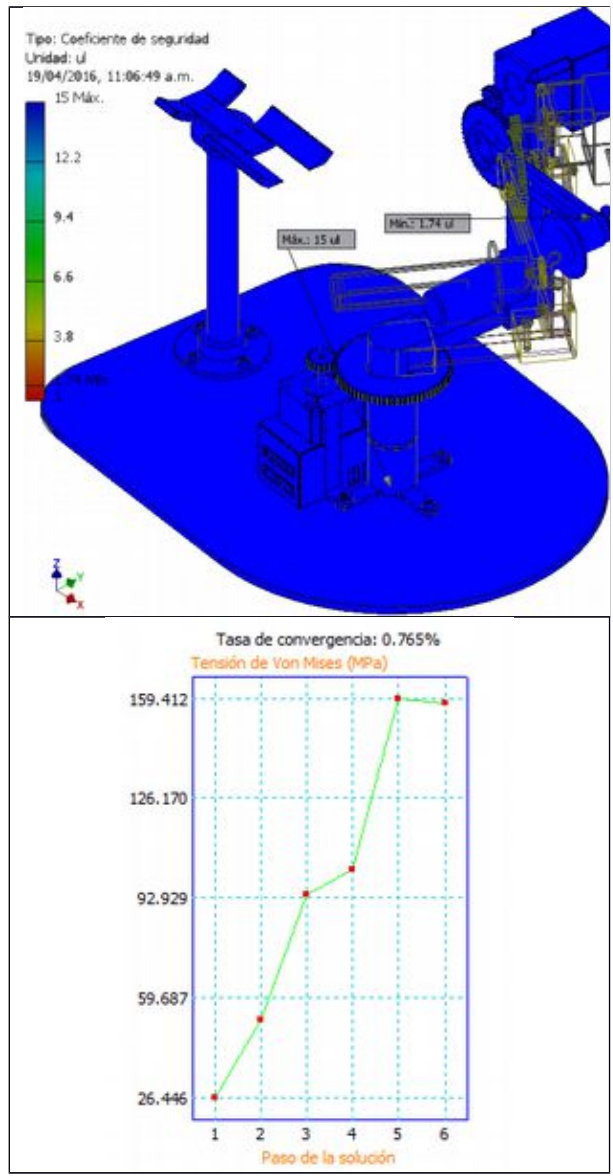

**Figura 8.** (a) Distribución de los factores de seguridad en el dispositivo de rehabilitación (b) Tasa de convergencia. Elaboración propia

En la figura 8, se observa que el factor de seguridad mínimo se encuentra por encima 1,74, lo cual indica que el equipo no presentará falla por deformación plástica. Adicionalmente se puede apreciar que la simulación presenta una tasa de convergencia del 0,765%, esto significa que la diferencia entre el esfuerzo o factor de seguridad calculado previamente al esfuerzo o factor de seguridad último presenta una variación del 0,765%, lo cual indica que se ha alcanzado la convergencia en los cálculos de los esfuerzos y factores de seguridad.

## 5. CONCLUSIONES.

En este trabajo se desarrolló el diseño de un dispositivo de rehabilitación pasiva capaz de registrar y ejecutar las rutinas de rehabilitación para la articulación de la muñeca a través del uso de un microcontrolador Arduino Uno R3, este dispositivo propuesto presentó características y funciones acordes con la mayoría de los requerimientos funcionales, tecnológicos, formales y ergonómicos planteados en los apartados anteriores, cubriendo así las necesidades de los pacientes.  Las formas de las geometrías propuestas en el dispositivo de rehabilitación son de fácil manufactura y ensamblaje, ya que se propuso el uso de perfiles comerciales y componentes comerciales de fácil acceso. El uso de materiales los asignados como el acero, aluminio 6061 y plástico ABS disponibles dentro de los almacenes del país hace que sea factible la construcción del dispositivo de rehabilitación. Mediante los análisis de ingeniería realizados, se obtuvieron parámetros de diseño cinemáticos como las velocidades angulares máximas instantánea con valores alrededores 432,42 °/s en el plano dorsal-palmar, esta velocidad de trabajo se encontró por debajo de la velocidad máxima angular de 3600 °/s alcanzadas por los moto-res paso-paso seleccionados STM17. Con respecto al estudio cinético se encontró que el valor máximo del momento impulsor fue de 184,5 N·mm localizándose en el plano de rotación flexión-extensión Dorsal-Palmar, igualmente este valor de par de trabajo también puede ser alcanzado por los motores paso-paso seleccionado de manera que permitirán ejecutar la rutina de rehabilitación según lo establecido por los médicos fisiatras del servicio de fisiatría. Del estudio estático de esfuerzos realizado al sistema estructural del equipo, se encontraron que las piezas críticas resultaron ser la arandela y el tornillo de ajuste mariposa alcanzado el valor de esfuerzo máximo de 159,41 MPa, y al determinar los Factores de Seguridad en el Equipo en general estos resultaron superiores a la unidad indicando que el equipo no presentara falla por deformación plástica. Al utilizar este dispositivo propuesto, el paciente presentara menos incomodidad durante el proceso de la ejecución de las rutinas de rehabilitación, ya que este fue diseñado con las características antropométricas de la población venezolana, y al ser un dispositivo de rehabilitación estacionario todo el peso de la mano del paciente se apoyará directamente sobre el dispositivo. En general se puede afirmar que el equipo desarrollado brindará una gran ayuda al médico fisiatra en la realización de las rutinas de rehabilitación, debido a que el médico especialista podrá supervisar y controlar diversas rutinas de rehabilitación con diferentes equipos a diferentes pacientes al mismo tiempo, evitándose así la gran fatiga al tener que realizar estas rutinas en jornadas repetitivas y prolongadas en el tiempo a todos los pacientes que atienden en el servicio de fisiatría.

## 6. AGRADECIMIENTOS.

## 7. REFERENCIAS.

[1] L. Almekinders, "Tendinitis and other chronic tendinopathies", Journal of American Academy of Orthopedic Surgery, vol 6, pp.157-164, 1998.

[2] B. Brotzman, "Handbook of Orthopaedic Rehabilitation", Editorial Mosby Inc., Estados Unidos de América, 1996.

[3] J. Oblack, I. Cikajlo, Z. Matja˘cic, " Universal Haptic Drive: A Robot for Arm and Wrist Rehabilitation". IEEE Transactions on Neural Systems and Rehabilitation Engi-neering, vol. 18, no. 3, pp. 293-302, Jun. 2010.

[4] V. Squeri, L. Masia, P. Giannoni, G. Sandini, P. Morasso, "Wrist Rehabilitation in chronic stroke patients by means of adaptive, progressive robot aided therapy". IEEE Transactions on Neural Systems and Rehabilitation Engineering, vol. 22, no. 2, pp. 312-25, Mar. 2014.

[5] L. Masia, H. Krebs, P. Cappa, N. Hogan, "Design and Characterization of Hand Module for Whole-Arm Rehabilitation Following Stroke". IEEE/ASME Transactions on Mechatronics, vol. 12, no. 4, pp. 399-407, Aug. 2007.

[6] R. Gutiérrez, et al, "Exoesqueleto Mecátronico para Rehabilitación Motora". Memorias del 8º Congreso Iberoa-mericano de Ingeniería Mecánica. Cusco, Perú. 2007.

[7] A. Gupta, M. O'Malley, V. Patoglu, C. Burgar, " Design, Control and Performance of Rice Wrist: A Force Feedback Wrist Exoskeleton for Rehabilitation and Training". The International Journal of Robotics Research, vol. 27, pp. 233, Feb. 2008.

[8] R. Song, K. Tong, X. Hu, W. Zhou, "Myoelectrically con-trolled wrist robot for stroke rehabilitation", Journal

Desarrollo de un Robot de Rehabilitación pasiva para la articulación de la muñeca mediante la implementación de un microcontrolador Arduino UNO of NeuroEngineering and Rehabilitation, vol 10, pp. 52, 2013.

[9] J. Buitrago, J. Lamprea, E. Caicedo, "Propuesta de un Framework multinivel para el diseño de laboratorio de acceso remoto". Rev.UIS.Ingenierías, vol.12, no.2, pp 35-45, 2013.

[10] A. Barraza et al, "Modelado dinámico del manipulador serial Mitsubishi Movemaster RV-M1 usando SolidWorks". Rev. UIS Ingenierias, vol. 15, no. 2, pp. 49-62, 2016.

[11] F. Aguayo F, V. Soltero, "Metodología del Diseño Industrial: un enfoque desde la Ingeniería Concurrente". Ed. Ra-Ma, Madrid. España, 2003.

[12] Fundación Centro de Estudios Sobre Crecimiento y Desarrollo de la Población Venezolana (FUNDACREDESA), "Tabla de peso, talla, circunferencia cefálica, longitudes de huesos de la población venezolana". Caracas, Venezuela, 1993.

[13] B. Le Veau B, "Biomecánica del movimiento Humano". Primera Edición, Editorial Trillas, México, 1991.

[14] J. Panero, M. Zelnik, "Las Dimensiones Humanas en los Espacios Interiores". Séptima Edición. Ediciones G. Giii, S.A., México. 1996.

[15] F. Beer, E. Johnston, J. DeWolf, "Mecánica de Materiales", cuarta edición, Mc Graw Hill, Mexico, 2007.

[16] J. Shigley, C. Mischke, "Diseño en ingeniería mecánica". Sexta edición, Mc Graw Hill, Mexico. 2002.

[17] D. M. Mercado, G.H. Murgas, J. R. Mckinley, J. D. González, "Una herramienta computacional didáctica para el análisis cinemático de mecanismos planos de cuatro barras". Revista UIS Ingenierías, vol. 14, no. 1, pp. 59-69, 2015.

[18] A. Barrios, "Fundamentos de Robótica, Segunda Edición, Editorial Mc Graw Hill, España. 2007.

[19] Robotzone LLC, "Catalogo de Productos Servocity". Recuperado de http//www.Servocity.com. 2016.

[20] M. Latarjet, A. Ruiz, "Anatomía humana". Editorial médica panamericana, Madrid, España. 1999.

[21] R.D. Chacón, L. J. Andueza, M.A. Díaz, J.A. Alvarado, "Analysis of stress due to contact between spur gears". En Advances in Computational Intelligence, Man-Machine Systems and Cybernetics. pp. 216-220, 2010.

[22] Phidgets DLL, "Manual de Selección de Controladores de Motores Paso-Paso Bipolar". USA. Recuperado de http//www.phidgets.com. 2016.

[23] J. Cazalilla., M. Vallés, A. Valera, V. Mata, M. Díaz-Rodríguez, "Hybrid force/position control for a 3-DOF 1T2R parallel robot: Implementation, simulations and experiments". Mechanics Based Design of Structures and Machines, vol. 44, no. 1-2, pp. 16-31, 2016.